\documentclass{article}

\PassOptionsToPackage{numbers, sort&compress}{natbib}

\usepackage[final]{neurips_2024_ml4ps}

\usepackage[utf8]{inputenc} 
\usepackage[T1]{fontenc}    
\usepackage{hyperref}       
\usepackage{url}            
\usepackage{booktabs}       
\usepackage{amsfonts}       
\usepackage{nicefrac}       
\usepackage{microtype}      
\usepackage{xcolor}         
\usepackage{graphicx}
\usepackage{subcaption}
\usepackage{amsmath}
\usepackage{enumitem}

\title{CODES: Benchmarking Coupled ODE Surrogates}

\author{%
  \href{mailto:robin.janssen@stud.uni-heidelberg.de}{Robin Janssen}, \href{mailto:immanuel.sulzer@stud.uni-heidelberg.de}{Immanuel Sulzer} and \href{mailto:tobias.buck@iwr.uni-heidelberg.de}{Tobias Buck} \\
  Interdisciplinary Center for Scientific Computing\\
  University of Heidelberg \\
}

\begin{document}

\maketitle

\begin{abstract}
We introduce \href{https://github.com/robin-janssen/CODES-Benchmark}{CODES}, a benchmark for comprehensive evaluation of surrogate architectures for coupled ODE systems. Besides standard metrics like mean squared error (MSE) and inference time, CODES provides insights into surrogate behaviour across multiple dimensions like interpolation, extrapolation, sparse data, uncertainty quantification, and gradient correlation. The benchmark emphasizes usability through features such as integrated parallel training, a web-based configuration generator, and pre-implemented baseline models and datasets. Extensive documentation ensures sustainability and provides the foundation for collaborative improvement. By offering a fair and multi-faceted comparison, CODES helps researchers select the most suitable surrogate for their specific dataset and application while deepening our understanding of surrogate learning behaviour.
\end{abstract}

\section{Introduction and Overview}

Coupled ordinary differential equations (ODEs) are ubiquitous in the natural sciences, and significant effort has gone into using machine learning models (surrogates) to replace the computationally expensive numerical methods required to solve such systems \cite{grassi_reducing_2022, branca_neural_2022, chemulator_2021, mace_2024, branca_emulating_2024}. Despite individual benchmarking efforts when presenting such new approaches, there is a lack of comprehensive and fair comparisons across different architectures. This makes it difficult to identify the best surrogate for a specific task, especially since the optimal choice can depend on both the dataset and the target application.

The CODES benchmark (\textbf{C}oupled \textbf{ODE} \textbf{S}urrogates) addresses this gap by providing a framework for evaluating surrogates in the context of coupled ODE systems. Beyond accuracy, training, inference time, and computational cost, other crucial factors differentiate architectures. As shown in \autoref{results}, surrogates vary in how they handle sparse data and perform in interpolation and extrapolation tasks. Some struggle more in highly dynamic regions, while others may yield unreliable uncertainty estimates, which could be problematic in hybrid approaches where numerical solvers are employed for predictions with low certainty. 
CODES highlights these differences, helping researchers and practitioners choose the best model for their task and deepening our understanding of whether surrogates truly capture the underlying dynamics of the data. The primary use cases include adding a new dataset and identifying the most suitable surrogate for it or adding a new surrogate to the benchmark and comparing its performance to the baseline models on existing datasets. 

CODES focuses on many essential aspects of benchmarking. Several \textbf{baseline models} and \textbf{datasets} are already implemented. Extensive \textbf{documentation}, featuring detailed doc-strings and type hints, ensures ease of use and project sustainability, supported by a dedicated \href{https://codes-docs.web.app}{documentation website} \cite{sulzer_codes_2024}. To guarantee \textbf{reproducibility}, all models are trained and evaluated in a fully seeded manner. Usability is crucial for broad adoption. Therefore, features such as a web-based config-file generator, parallel training support, toggleable comparison modalities, and automated evaluation through tables and plots have been integrated. CODES is intended to grow as a  \textbf{collaborative} project, with clear instructions on adding new datasets and surrogate models, encouraging community contributions.

\section{Structure of the Benchmark}
\label{structure}

CODES currently includes four surrogate architectures and five datasets. After configuration, the models are first trained on the training set and then evaluated on the test set. The code is available on GitHub \cite{codes_repo}.

\paragraph{Datasets \& Surrogates.}

\begin{figure}[ht!]
  \centering
  \includegraphics[width=1\textwidth]{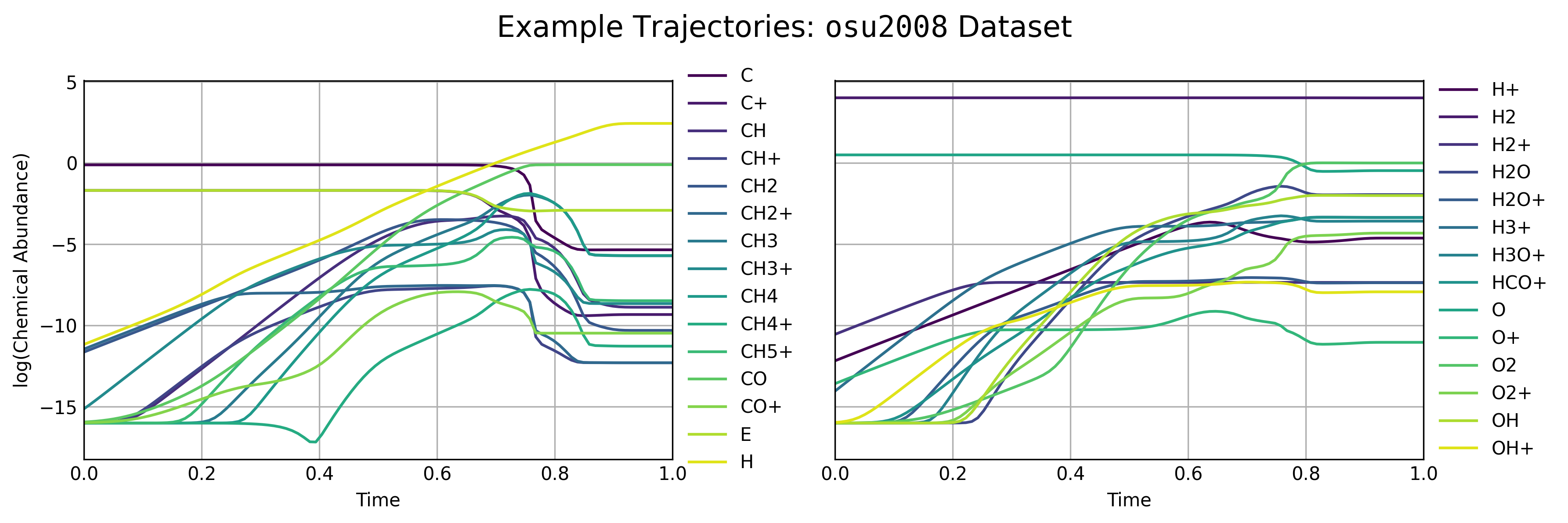}
  \caption{Example trajectories of a datapoint (29 chemicals, 100 timesteps) in the \texttt{osu2008} dataset.}
  \label{fig:trajectories}
\end{figure}

CODES is largely inspired by astrochemistry, hence the two most complex datasets come from this field: \verb|osu2008| was adapted from Grassi et al. \cite{grassi_reducing_2022} (an exemplary plot is shown in \autoref{fig:trajectories}) while \verb|branca24| was kindly provided by Branca \& Pallottini from their recent paper \cite{branca_emulating_2024}. The remaining three datasets (\verb|lotka_volterra|, \verb|simple_ode|, and \verb|simple_reaction|) are lower-dimensional baselines created by the authors of this paper. All datasets are hosted on Zenodo and automatically downloaded as needed.  \\
The following architectures are currently implemented: 
\begin{itemize}[left=10pt, itemsep=1pt, topsep=0pt]
\item \verb|FullyConnected| (\verb|FCNN|). A standard fully-connected neural network.
\item \verb|MultiONet| (\verb|MON|). An adaptation of DeepONet \cite{lu_deeponet_2019} for multiple outputs as proposed in \cite{Lu_2022}.
\item \verb|LatentNeuralODE| (\verb|LNODE|). An autoencoder with a latent space neural ODE, following \cite{sulzer_speeding_2023}
\item \verb|LatentPoly| (\verb|LP|). An autoencoder with a learnable latent space polynomial, also following \cite{sulzer_speeding_2023}.
\end{itemize}

 All models receive as input the abundances at $t=0$ and then predict abundances for some later time $t>0$. Time is treated either as an additional input (\verb|FCNN|, \verb|MON|) or as a parameter that determines the length of the latent-space evolution (\verb|LNODE|, \verb|LP|).  Detailed explanations for all architectures and datasets are available in \autoref{appendix:surrogates} and \autoref{appendix:datasets} as well as on the documentation website \cite{sulzer_codes_2024}.

\paragraph{Modalities.}

In addition to training a main model on the full dataset, the following modalities can be toggled to train additional models:

\begin{itemize}[left=10pt, itemsep=1pt, topsep=0pt]
\item \verb|Interpolation| skips timesteps (or generally the independent variable) in a specified interval. 
\item \verb|Extrapolation| trains only on timesteps below a set threshold. 
\item \verb|Sparse| thins out the training data by a specified factor, using only every $n$-th sample. 
\item \verb|Batch| trains the models with a user-defined batch size. 
\item \verb|Uncertainty| trains $N$ additional models with different random initializations and shuffled datasets to create a DeepEnsemble \cite{deepensemble} for uncertainty estimation.
\end{itemize}

By setting multiple values (i.e., multiple intervals, thresholds, factors, or batch sizes), users can gain insights into and compare surrogate behaviour, answering questions such as: Which model handles sparse data best? Which performs the best interpolation?
For evaluation, users can configure additional output such as inference time, memory usage, loss curves, and different correlations. A configuration tool on the website \cite{sulzer_codes_2024} helps users generate the \texttt{config.yaml}.

\newpage

\paragraph{Training \& Benchmarking.}

Training is the most time-intensive step in the benchmark, as for each setting an additional model is trained on the corresponding subset of training data (e.g., 96 models were trained for the evaluations presented in \autoref{results}). 
To accelerate this process, training is parallelized across all specified devices using a task-list approach. Progress bars show the status of active jobs. Custom hyperparameters can be specified for each surrogate, and predefined optimal hyperparameters are available for the provided datasets.
After the training run finishes, evaluation results are generated in the form of metrics and plots. These include individual (per-surrogate) and comparative results. Currently, up to 18 individual plots per surrogate and 12 comparative plots are provided. Metrics include standard measures like mean-squared error (MSE), mean absolute error (MAE), mean relative error (MRE), and inference time. Additionally, there are more advanced metrics, such as the Pearson correlation coefficient (PCC) between predictive uncertainty (via DeepEnsemble) and prediction error (PE), or between data gradients and prediction error. See \autoref{tab:model-comparison} for a selection of the available metrics.

\section{Results}
\label{results}

The following results represent an exemplary evaluation of the benchmark. Models were trained on the \verb|osu2008| dataset, generated from a chemical reaction network with 224 reactions among 29 chemical species, simulated at constant temperature over 100 time steps. The dataset contains 1000 data points, split into 75\% training, 5\% validation, and 20\% testing. Models were trained on NVIDIA TITAN Xp GPUs and evaluated on an NVIDIA GeForce RTX 2080 Ti. Optimal configurations for each model were determined using \href{https://optuna.org}{Optuna} \cite{optuna_2019} and are detailed in \autoref{appendix:surrogates}. While the results are specific to this dataset, broader patterns can be observed.
Given the benchmark’s comprehensive nature, the following results are just a subset of the analyses and insights that can be drawn. Additional plots can be found in \autoref{appendix:plots} or on the documentation website \cite{sulzer_codes_2024}.

\paragraph{Performance.}

\begin{figure}
  \centering
  \includegraphics[width=1\textwidth]{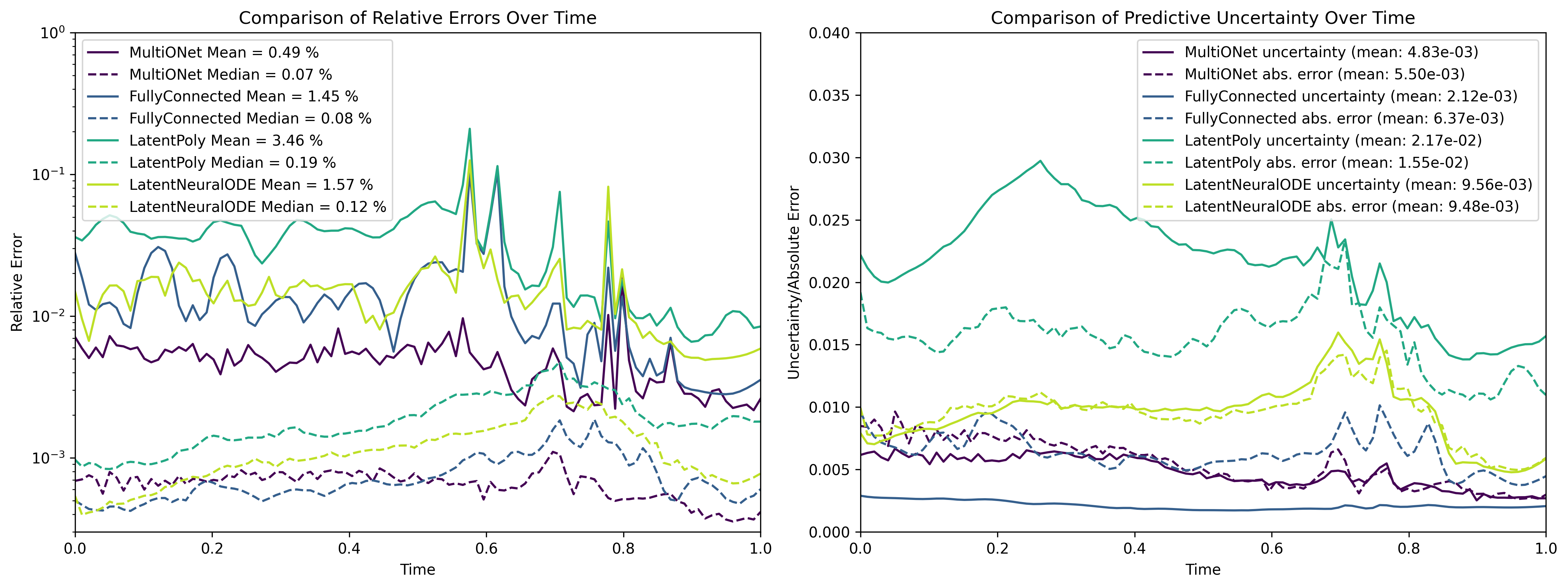}
  \caption{Quantities over time per model, averaged across the test set. \textbf{Left}:  Mean and median relative error. \textbf{Right}: Predictive uncertainty (DeepEnsemble, $n=5, 1\sigma$) and mean absolute error.}
  \label{fig:rel_errors_and_uq}
\end{figure}

The main benchmark metrics, along with additional model parameters, are shown in \autoref{tab:model-comparison}. Despite significant differences between models, all show impressive performance — even the slowest model, \verb|LNODE|, achieves speedups of at least three orders of magnitude compared to numerical methods (as discussed by Sulzer \& Buck in \cite{sulzer_speeding_2023}) and all models exhibit mean relative errors of below 4\%. \verb|MON| performs best on accuracy metrics such as MSE, MAE and MRE (followed closely by \verb|FCNN|) , while \verb|LP| has the fastest inference time, likely due to the computational simplicity of evaluating polynomials with fixed coefficients. \autoref{fig:rel_errors_and_uq} displays the mean and median error of each surrogate over time. The spikes in the second half correspond to more dynamic sections of the data, with which some surrogates struggle more than others (cf. the section on UQ and correlations).

\begin{table}[ht!]
  \caption{Benchmark Results for the Surrogate Models}
  \label{tab:model-comparison}
  \centering
  \begin{tabular}{lcccc}
    \toprule
    Metric                & MultiONet         & FullyConnected    & LatentPoly               & LatentNeuralODE    \\
    \midrule
    MSE                   & \textbf{5.89e-05} & 8.57e-05          & 4.50e-04                 & 1.86e-04           \\
    MAE                   & \textbf{5.50e-03} & 6.37e-03          & 1.55e-02                 & 9.48e-03           \\
    MRE                   & \textbf{0.49\%}   & 1.45\%            & 3.46\%                   & 1.57\%             \\
    Inference Time (ms)   & 23.37 ± 0.10      & 3.38 ± 0.10       & \textbf{0.55 ± 0.06}     & 37.76 ± 0.84       \\
    Mean uncertainty      & 4.83e-03            & 2.12e-03            & 2.17e-02                   & 9.56e-03             \\
    PCC UQ                & 0.3551            & 0.1909            & \textbf{0.4178}          & 0.4166             \\
    PCC Gradient          & 0.2606            & 0.3054            & 0.3076                   & \textbf{0.5006}    \\
    Epochs                & 1000              & 1000              & 15000                    & 10000              \\
    Train Time (hh:mm:ss) & 02:51:36          & 01:51:45          & \textbf{00:54:09}        & 01:49:30           \\
    \# Trainable Params   & 558970            & 184429            & 62864                    & 72368              \\
    \bottomrule
  \end{tabular}
\end{table}

\newpage

\paragraph{Interpolation, Extrapolation, and Sparsity.}

\begin{figure}
  \centering
  \includegraphics[width=1\textwidth]{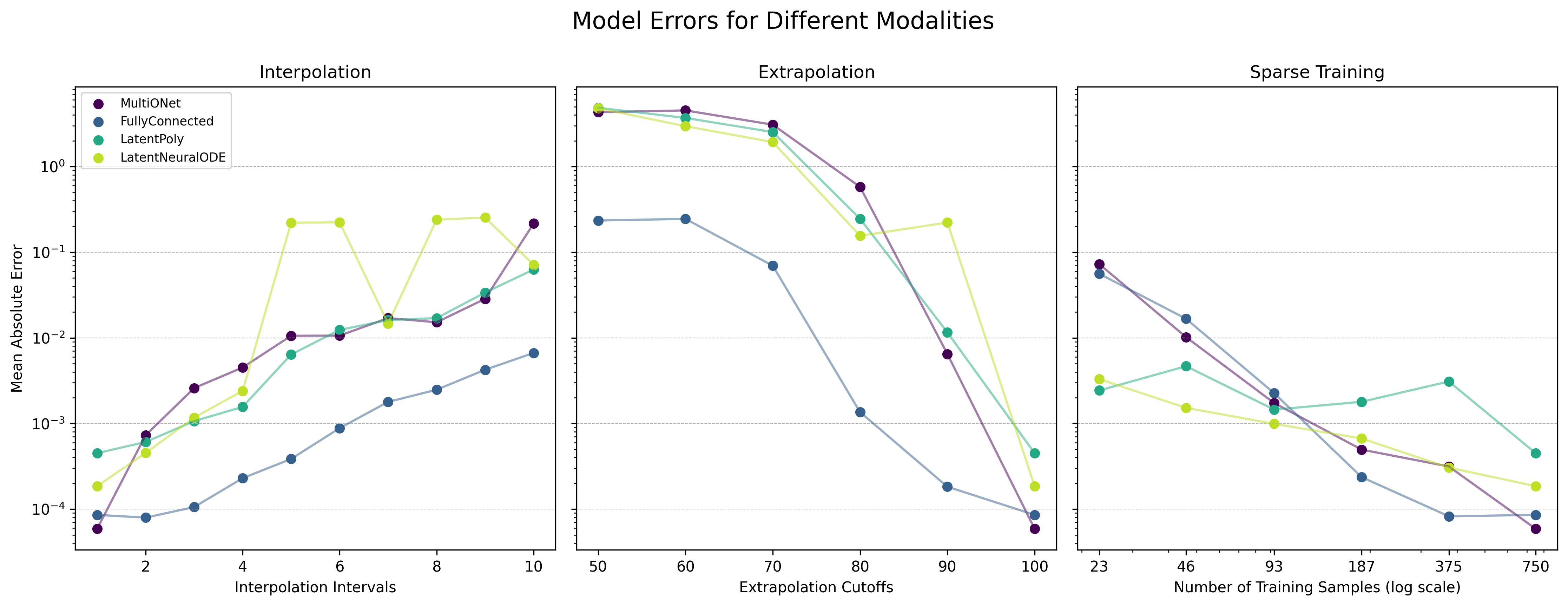}
  \caption{Model errors for the modalities \texttt{interpolation}, \texttt{extrapolation} and \texttt{sparse}.}
  \label{fig:generalisation}
\end{figure}

The differences in inductive bias between surrogates become more apparent under challenging conditions. \autoref{fig:generalisation} shows the model MAE across three different benchmark modalities. In \texttt{interpolation} and \texttt{extrapolation}, \verb|FCNN| outperforms all other models. The spikes observed in \verb|LNODE| may result from intervals where the remaining timesteps are not representative of the full dataset, a challenge the other models seem to handle more effectively. In the \texttt{sparse} data setting, the autoencoder-type models, \verb|LNODE| and \verb|LP|, outperform the fully-connected models, \verb|FCNN| and \verb|MON|, for low numbers of training samples ($\leq 93$),  suggesting greater robustness to settings with reduced information. This might be explained by the fact that the former involve a temporal evolution, which more closely mirrors the problem structure, while the latter treat time merely as an additional input.

\paragraph{Uncertainty Quantification \& Correlations.}

Comparing the predictive uncertainty (the $1 \sigma$ interval from DeepEnsemble predictions) with the mean absolute error over time in \autoref{fig:rel_errors_and_uq} allows for insights into how well each surrogate architecture estimates its own prediction errors. DeepEnsemble seems to provide a generally reliable uncertainty estimate, particularly for \verb|LNODE| and \verb|MON|. 
This is also confirmed by the close coincidence of the MAE and mean uncertainty metrics in \autoref{tab:model-comparison}. \verb|FCNN| tends to be overconfident, underestimating the magnitude of its errors, while \verb|LP| slightly tends towards the opposite behavior. 
It could be argued that the magnitude of an uncertainty estimate is less important than its correlation to the errors, since the estimate could just be rescaled appropriately. Following this argument, we added the Pearson correlation coefficient (PCC) between the predicted uncertainty and the prediction error (calculated across all predictions in the test set) to the benchmark. This metric is lowest for \verb|FCNN|, confirming the apparent overconfidence of this architecture.
Additionally, we integrated the PCC between data gradients and prediction error. This metric is highest for \verb|LNODE|, evidencing a proclivity towards higher errors at points with larger gradients. This provides an additional indication to our earlier impression that this architecture struggles more with non-smooth data than other architectures.

\section{Discussion}
\label{sec:discussion}

\paragraph{Summary.}

The CODES benchmark offers diverse insights into the behavior of surrogates for coupled ODE systems. Beyond identifying the fastest or most accurate architecture, its extensive plots and metrics help determine the most suitable surrogate for a specific dataset and application, which may require characteristics such as robustness to sparse data, reliable interpolation, or precise uncertainty estimation. The benchmark is modular, with a focus on usability, reproducibility, and collaboration. A comprehensive documentation website, including a web-based configuration tool, was created to support this effort \cite{sulzer_codes_2024}.

\paragraph{Limitations.}

Ensuring fair and representative comparisons can be challenging. CODES addresses this by seeding and training all models consistently within the benchmark. The limitations of the benchmark can roughly be divided into two categories: the influence of architectural differences on comparability and the data-dependency of the results. \\
For the results in \autoref{results}, models were compared by training each until its loss trajectory stabilized, which is an effective but potentially imprecise approach. Alternatives such as training for a fixed number of epochs (implemented) or equal training time (not yet implemented) could also be argued for, and would significantly affect benchmark outcomes. Model performance heavily depends on hyperparameters, which were optimized via Optuna. This process is not yet integrated into the benchmark and must be performed manually; it may bias some models towards higher complexity, sacrificing speed. Architectural differences further complicate the comparison of the inference time. For example, \verb|LNODE| is more efficient when predicting entire trajectories at once rather than generating individual predictions step-by-step, potentially skewing the comparison. We adopted the trajectory-based approach, but its representativeness depends on the target application. \\
In principle, training surrogates to model an ODE is an infinite-data problem because the ground truth data is generated with a numerical solver from randomly sampled initial conditions. In addition to allowing for the generation of arbitrary amounts of training data, this problem structure also invites online or active learning approaches. However, the benchmark currently only supports finite datasets provided in hdf5 format.  Finally, the benchmarking results are always dataset-specific. This implies that a broad range of datasets is needed to thoroughly assess surrogate performance and necessitates further insights into each dataset to inform comparisons between them. Additionally, findings are only reliable if the distribution of the training data matches that of the target application, emphasizing the need for representative training data. \\

\paragraph{Outlook.}
CODES enables comprehensive comparisons of new and existing surrogate architectures and helps users select the ideal model for their specific context, ensuring optimal performance in target applications. By improving surrogate selection, CODES contributes to faster and more reliable simulation frameworks, enabling higher resolution or more complex applications. Future updates will include integrated hyperparameter tuning, dataset-specific metrics and visualizations, further exploration of inter- and extrapolation capabilities, additional baseline models, and potentially support for online training.  A key focus will be adding new datasets and surrogates, for which we warmly invite community contributions that will be credited appropriately.

\begin{ack}
This work is funded by the Carl-Zeiss-Stiftung through the NEXUS program and supported by Deutsche Forschungsgemeinschaft (DFG, German Research Foundation) under Germany’s Excellence Strategy EXC-2181/1 - 390900948 (the Heidelberg STRUCTURES Cluster of Excellence) via the exploratory project EP10.2.
\end{ack}

\newpage

\bibliography{bibliography_new}

\newpage

\appendix

\section{Surrogate Architectures}
\label{appendix:surrogates}

\paragraph{Architectures.}Four architectures are currently implemented in the benchmark. Let $N_q$ denote the number of quantities in a dataset (e.g., $N_q = 29$ for the \verb|osu2008| dataset).
\begin{itemize}
    \item \verb|FullyConnected| is a standard fully-connected neural network. It has $N_q + 1$ inputs since the time $t$ of prediction is given as extra input and $N_q$ outputs. 
    \item \verb|MultiONet| follows the DeepONet architecture first described in \cite{lu_deeponet_2019}. It features two fully-connected neural networks, the branch and trunk networks. The branch network receives as inputs the $N_q$ initial conditions, the trunk network receives as a single input the time $t$ of prediction. In the original architecture, the network has only a single output which is generated by taking the scalar product of the latent (output) vectors of branch and trunk. As proposed in \cite{Lu_2022}, we adapt this architecture to produce multiple outputs in a single forward pass by splitting the latent vectors of both networks into $N_q$ parts and taking the scalar product of corresponding parts of branch and trunk network. 
    \item \verb|LatentNeuralODE| features an autoencoder which is essentially a fully-connected neural network with a smaller output size $L_q$ than its input size $N_q$ which encodes an initial set of abundances into a latent space. There, the encoded initial condition is used as the initial state for a numerical integration of a second neural network which generates latent space trajectories through time (Neural ODE). Finally, the entire latent space trajectory is fed to a third neural network (the decoder) with input size $L_q$ and output size $N_q$ which outputs the predicted time series.
    \item \verb|LatentPoly| uses a similar encoder/decoder setup as \verb|LatentNeuralODE|, but instead of a Neural ODE the initial condition in latent space is added to a learnable polynomial which is evaluated at the time points of interest. The resulting latent space trajectory is again decoded back to the original dimensionality $N_q$.
\end{itemize}

\paragraph{Hyperparameters.}
All architectures used the Adam optimiser, the number of epochs are given in \autoref{tab:model-comparison}. The results in \autoref{results} were achieved using the following hyperparameters:
\begin{itemize}
    \item \verb|FullyConnected|: The architecture consists of two hidden layers with 400 neurons per layer, uses \texttt{tanh} activation functions and was trained with a learning rate of $1.5\cdot10^{-5}$.
    \item \verb|MultiONet|: The branch net features four hidden layers, the trunk net seven hidden layers, both with 150 neurons per layer. The output layer features 40 neurons per quantity for a total of 1160 outputs per network (which are then multiplied in the split scalar product described in \autoref{appendix:surrogates}). The training was performed with \texttt{LeakyReLU} activation and a learning rate of $5\cdot10^{-4}$.
    \item \verb|LatentNeuralODE|: The encoder features hidden layers of the sizes 184, 92 and 46, the final latent space layer has nine outputs. The decoder has the same architecture, but inverted in order. The training was performed with a learning rate of $5\cdot10^{-3}$, which was scheduled to decay to $10^{-5}$ over the course of the run.  The activaton function of the neural ODE is \verb|Softplus|, the encoder uses \verb|ReLU|. The numerical integration scheme used is \verb|Tsit5|.
    \item \verb|LatentPoly|: The encoder features hidden layers of the sizes 200, 100 and 50, the final latent space layer has five outputs and uses \verb|ReLU| activation. The decoder has the same architecture, but inverted in order. The latent-space polynomial has degree six, and the training was performed with a learning rate of $2\cdot10^{-3}$.
\end{itemize}
These settings are also stored in a surrogate configuration file for the \verb|osu2008| dataset and will be used automatically for training unless disabled in the configuration file. To reproduce the results of the benchmark, use \verb|seed: 42|.

\section{Datasets}
\label{appendix:datasets}

CODES currently comprises five datasets, two from the context of astrochemistry and three lower-dimensional baseline datasets. All datasets are hosted on Zenodo and are automatically downloaded when required. The number of samples, timesteps and quantities is displayed in \autoref{tab:datasets}. They are stored as \verb|hdf5| files containing the following data: 
\begin{itemize}
    \item Training, validation and test data stored as three separate numpy arrays. Training data is used to train models, validation data is used to compute intermediate accuracy and loss values (and for Optuna optimisation) while test data is used for the evaluation after training the models.
    \item The number of training, validation and test samples as well as the number of timesteps and evolving quantities (e.g. chemicals), all stored as integers.
    \item Optionally, timesteps and labels for the quantities can be included. The former are stored as an array of floating-point numbers, the latter as an array of strings. When provided, they will be used in plots in the benchmark.
\end{itemize}

The following datasets are currently provided in the benchmark:
\begin{itemize}
    \item \href{https://zenodo.org/records/13359976}{\texttt{osu2008}} was generated  from a chemical reaction network with 224 reactions among 29 chemical species, simulated at constant temperature over 100 time steps. Exemplary trajectories of this dataset can be found in \autoref{fig:trajectories}. For more details, refer to \cite{sulzer_speeding_2023}.
    \item \href{https://zenodo.org/records/13624794}{\texttt{branca24}} is by far the largest dataset in the benchmark. It describes the evolution of ten quantities over 16 timesteps. The dataset is a 1 \% subset of the dataset used by Branca \& Pallottini in \cite{branca_emulating_2024}, obtained by shuffling and slicing the original dataset.
    \item \href{https://zenodo.org/records/13624788}{\texttt{lotka\_volterra}} A six species lotka-volterra system, featuring three predator and three prey species, integrated over 100 time steps. The corresponding ODE system has the following equations

    \[
    \begin{aligned}
        \frac{dp_1}{dt} &= 0.5 p_1 - 0.02 p_1 q_1 - 0.01 p_1 q_2, \\
        \frac{dp_2}{dt} &= 0.6 p_2 - 0.03 p_2 q_1 - 0.015 p_2 q_3, \\
        \frac{dp_3}{dt} &= 0.4 p_3 - 0.01 p_3 q_2 - 0.025 p_3 q_3, \\
        \frac{dq_1}{dt} &= -0.1 q_1 + 0.005 p_1 q_1 + 0.007 p_2 q_1, \\
        \frac{dq_2}{dt} &= -0.08 q_2 + 0.006 p_1 q_2 + 0.009 p_3 q_2, \\
        \frac{dq_3}{dt} &= -0.12 q_3 + 0.008 p_2 q_3 + 0.01 p_3 q_3.
    \end{aligned}
    \]

    \item \href{https://zenodo.org/records/13624783}{\texttt{simple\_ode}} This is a simple, five dimensional ODE system integrated over 100 time steps.
    \[
    \begin{aligned}
    \frac{dn_0}{dt} &= -0.8 n_0 - 0.2 n_0 n_2, \\
    \frac{dn_1}{dt} &= 0.8 n_0 - 0.5 n_1 + 0.4 n_0 n_2, \\
    \frac{dn_2}{dt} &= 0.5 n_1 - 0.2 n_0 n_2, \\
    \frac{dn_3}{dt} &= 0.2 n_0 + 0.625 n_1, \\
    \frac{dn_4}{dt} &= 1.6 n_0n_2 - 0.5 n_0 n_2.
\end{aligned}
    \]
    \item \href{https://zenodo.org/records/13624781}{simple\_reaction} is a six dimensional ODE system which resembles a simple chemical rate equation ODE system, integrated over 100 time steps.

\[
\begin{aligned}
    \frac{ds_1}{dt} &= -0.1 s_1 + 0.1 s_2, \\
    \frac{ds_2}{dt} &= 0.1 s_1 - 0.15 s_2 + 0.05 s_3, \\
    \frac{ds_3}{dt} &= 0.15 s_2 - 0.1 s_3 + 0.03 s_4, \\
    \frac{ds_4}{dt} &= 0.1 s_3 - 0.07 s_4 + 0.01 s_5, \\
    \frac{ds_5}{dt} &= 0.07 s_4 - 0.05 s_5, \\
    \frac{ds_6}{dt} &= 0.05 s_5.
\end{aligned}
\]

\end{itemize}

\begin{table}[ht!]
\centering
\caption{Dataset Size Overview}
\label{tab:datasets}
\begin{tabular}{lcccc}
\toprule
\textbf{Dataset} & \textbf{Samples} & \textbf{Train/Val/Test Split} & \textbf{Timesteps} & \textbf{Quantities} \\
\midrule
\verb|osu2008|          & 1000    & 750 / 50 / 200   & 100  & 29 \\
\verb|branca24|         & 671089  & 503316 / 33554 / 134219   & 16   & 10 \\
\verb|lotka_volterra|   & 700     & 500 / 50 / 150   & 100  & 6  \\
\verb|simple_ode|       & 700     & 500 / 50 / 150   & 100  & 5  \\
\verb|simple_reaction|  & 700     & 500 / 50 / 150   & 100  & 6  \\
\bottomrule
\end{tabular}
\end{table}

\section{Benchmark Plots}
\label{appendix:plots}

\begin{figure}[ht!]
  \centering
  \includegraphics[width=1\textwidth]{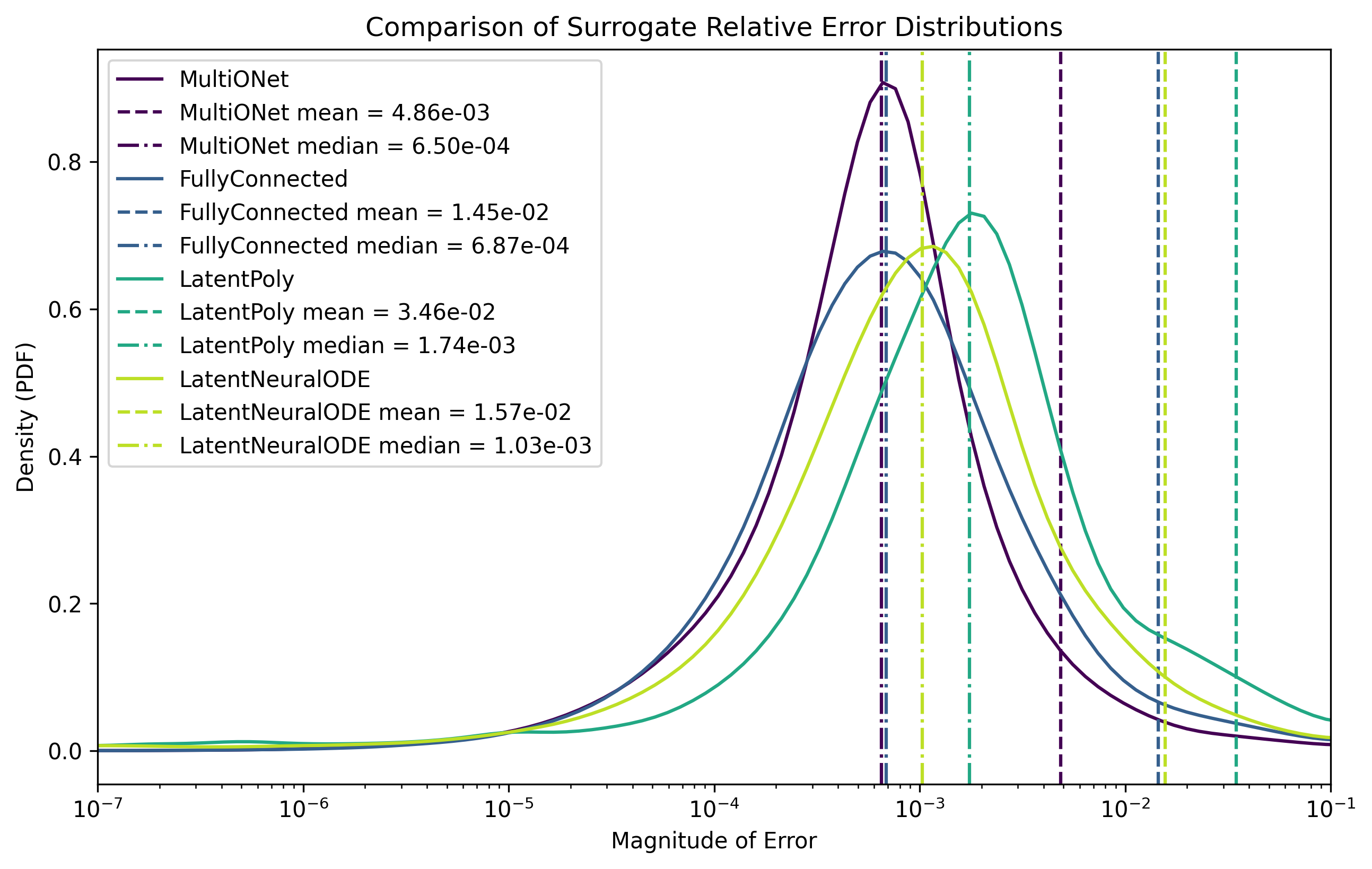}
  \caption{Smoothed histogram plots of the distribution of the relative errors per model alongside their mean and median relative errors.}
  \label{fig:model_error_dists}
\end{figure}

This section contains a selection of additional plots generated during a benchmark run. Note that this is not an exhaustive list, but rather an overview of some additional capabilities of the benchmark. All plots below relate to results obtained on the \verb|osu2008| dataset during the same benchmark run detailed in \autoref{results} with hyperparameters described in \autoref{appendix:surrogates}.

\begin{figure}[ht!]
  \centering
  \includegraphics[width=1\textwidth]{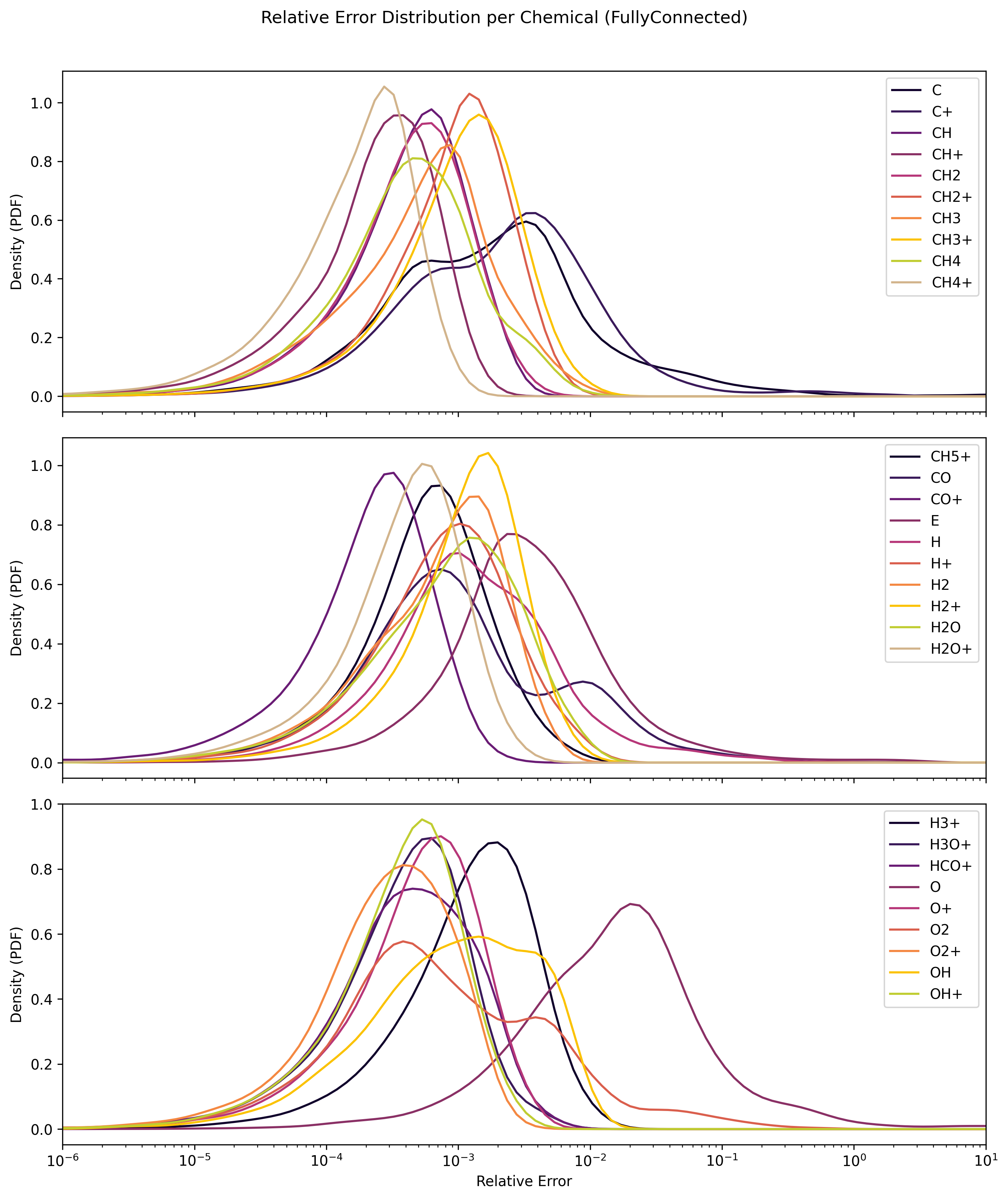}
  \caption{Smoothed histogram plots of the relative error distribution of \texttt{FullyConnected} for each quantity (chemical) in the \texttt{osu2008} dataset.}
  \label{fig:quantity_error_dists}
\end{figure}

\autoref{fig:model_error_dists} displays smoothed histogram plots of the relative error distribution of each model on the entire test set. The vertical lines indicate the mean and median error of the respective model. The superior accuracy of \verb|FullyConnected| and \verb|MultiONet| described in \autoref{results} is confirmed here, the peak of the distribution as well as the mean and median lines are lower than those of \verb|LatentNeuralODE| and \verb|LatentPoly|.

\begin{figure}[ht!]
  \centering
  \includegraphics[width=1\textwidth]{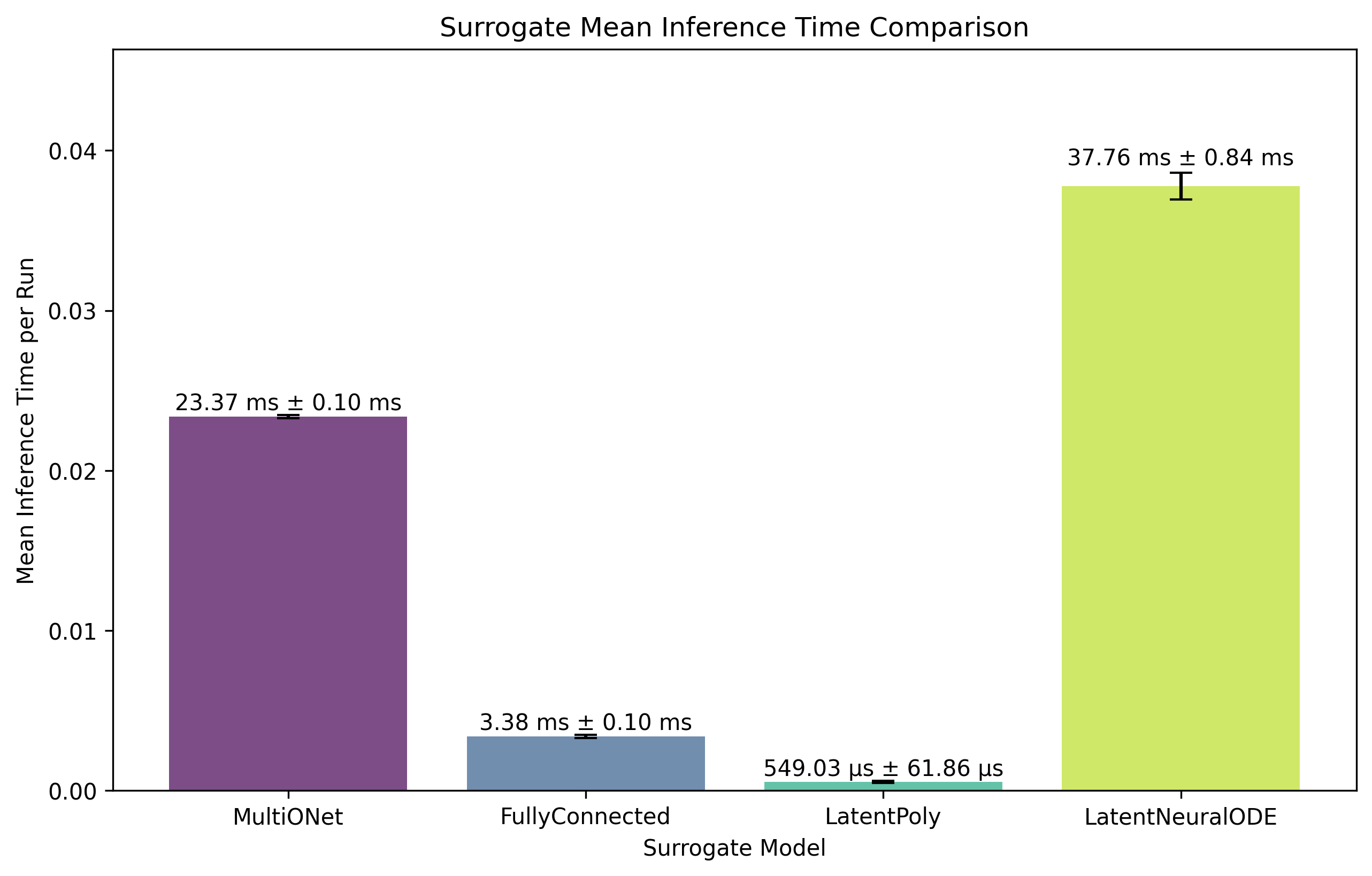}
  \caption{Bar plot of the inference times of all models for predicting the entire test set once.}
  \label{fig:inference_times}
\end{figure}

\begin{figure}[ht!]
  \centering
  \includegraphics[width=1\textwidth]{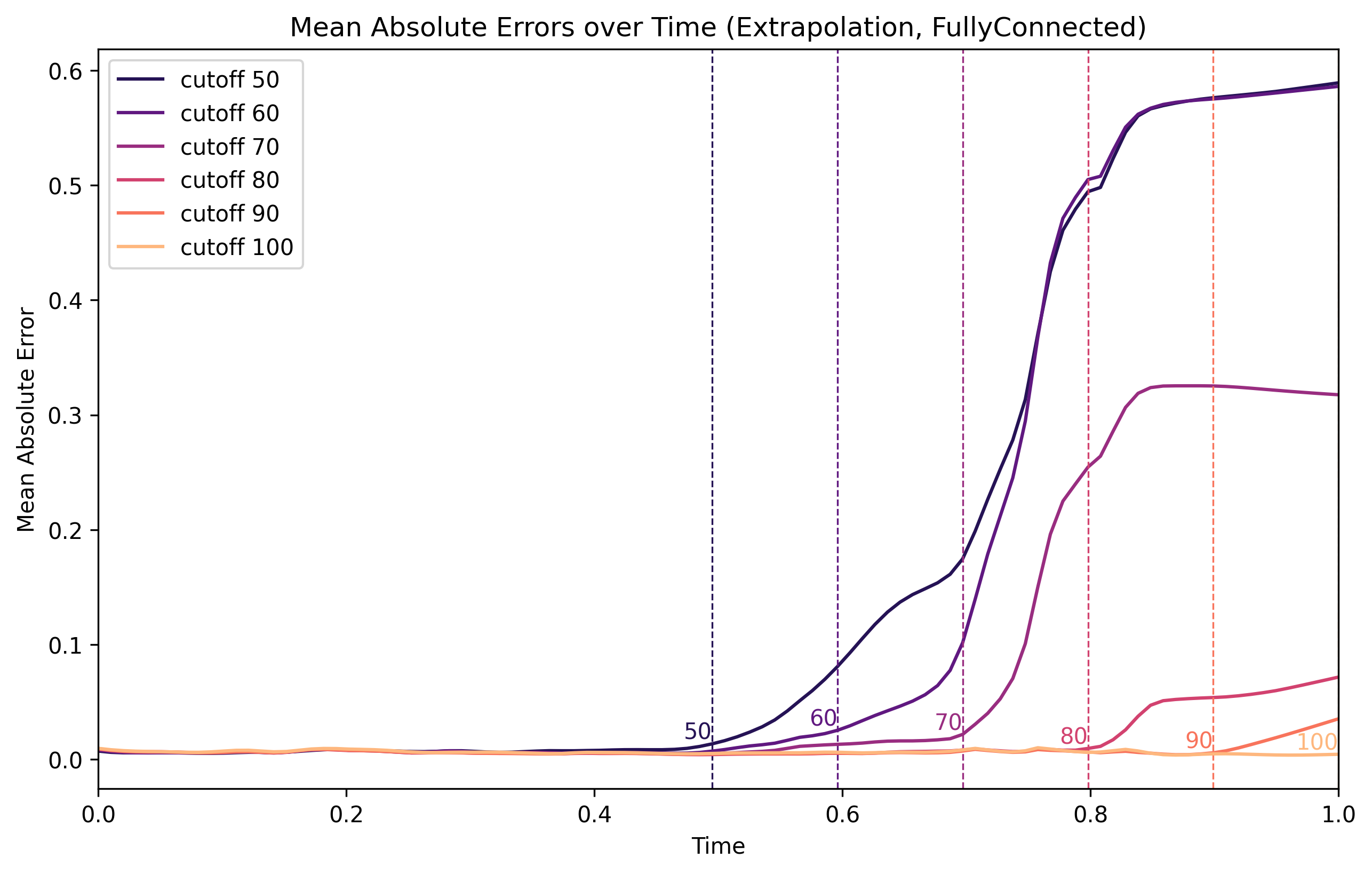}
  \caption{Mean absolute errors over time of \texttt{FullyConnected} for different \texttt{extrapolation} cutoffs.}
  \label{fig:extrapolation}
\end{figure}

\autoref{fig:quantity_error_dists} shows an exemplary relative error distribution per quantity for \verb|FullyConnected|. The benchmark generates a corresponding plot for each surrogate. These plots are useful for getting a feeling for how the surrogate performs on each quantity in the dataset. Notably, some quantities skew quite far towards large relative errors (O), small relative errors (CH4+) or exhibit two distinct peaks (CO).

\autoref{fig:inference_times} displays a bar plot of the inference times, which are measured as the time required for one full prediction of the test set (200 datapoints, 100 timesteps each). The errors were calculated by timing this duration five times to obtain a mean and standard deviation. Fluctuations in the inference times are very low and should result mostly from slight differences in the computational load of the processors or GPU during inference.

\autoref{fig:extrapolation} is another model-specific plot, showing the mean absolute error over time for \verb|FullyConnected| in the \verb|extrapolation| modality for the different time cutoffs the model was trained with. One can clearly discern how error increases over time once the model predicts timesteps that were not included in the training set. Similar plots are also generated for the modalities \verb|interpolation|, \verb|sparse| and \verb|UQ| for all surrogatess.

\begin{figure}[ht!]
  \centering
  \begin{subfigure}[b]{0.49\textwidth}
    \centering
    \includegraphics[width=\textwidth]{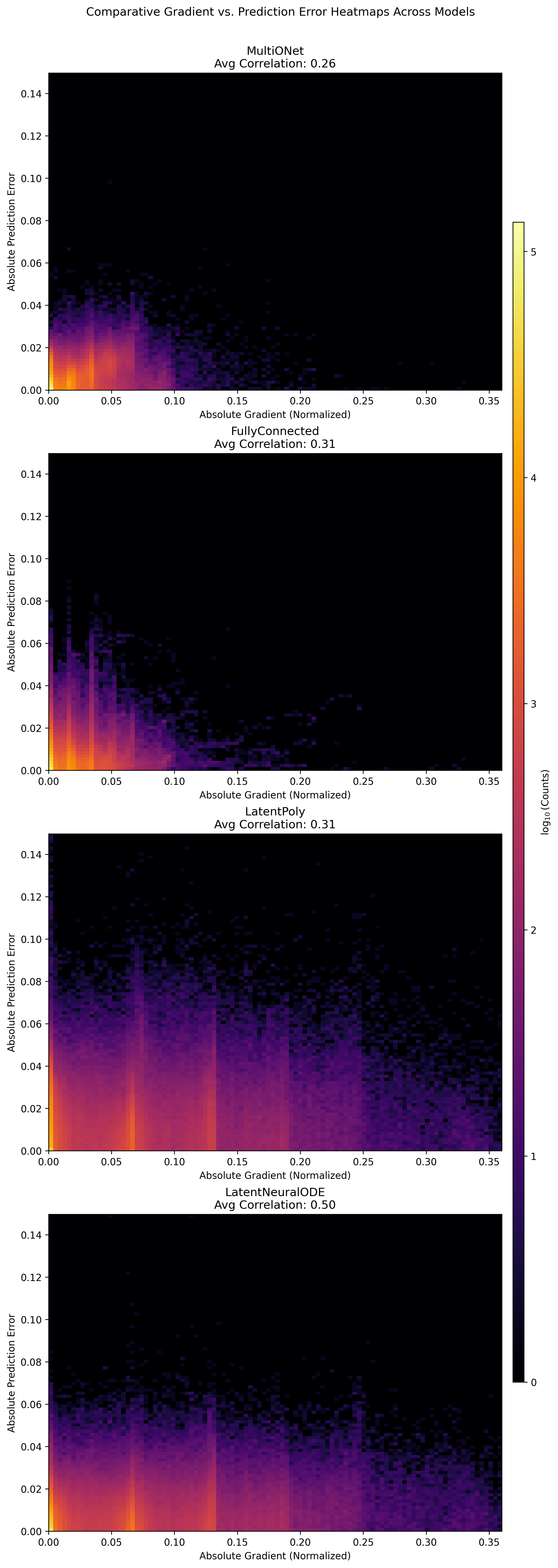}
    \caption{Gradient-error correlation heatmaps.}
    \label{fig:gradients_heatmap}
  \end{subfigure}
  \hfill
  \begin{subfigure}[b]{0.49\textwidth}
    \centering
    \includegraphics[width=\textwidth]{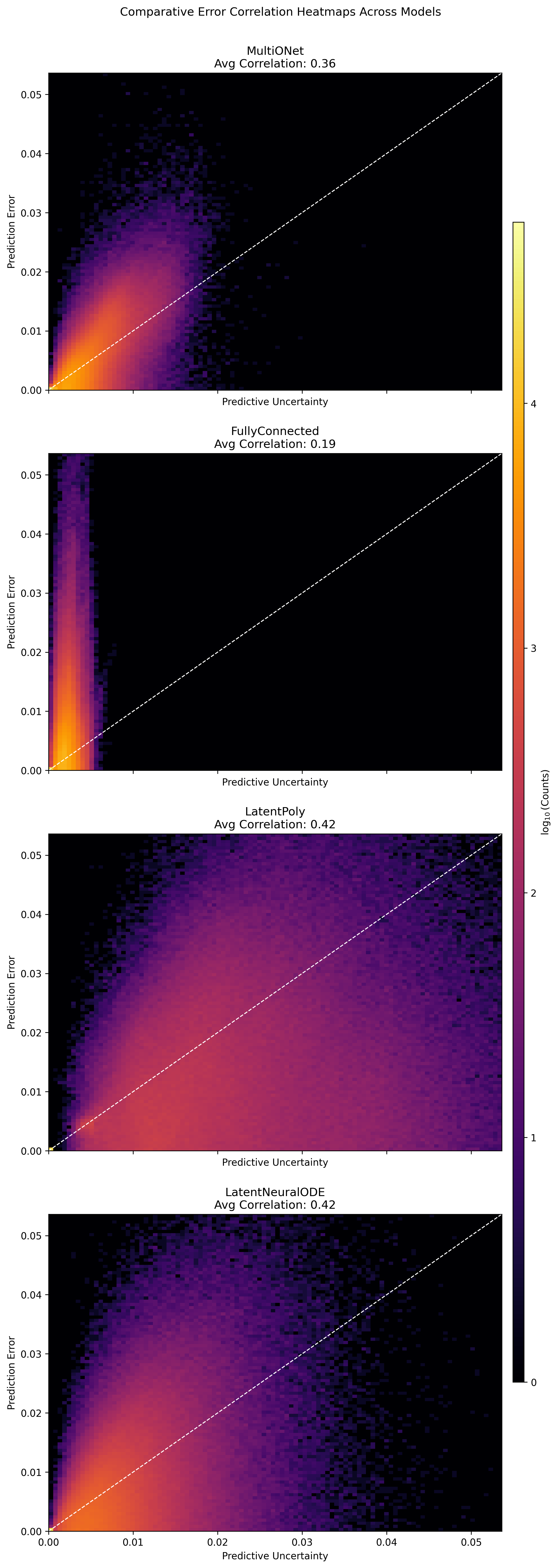}
    \caption{Uncertainty-error correlation comparison.}
    \label{fig:uq_heatmap}
  \end{subfigure}
  \caption{Heatmaps visualising the correlations between \textbf{a)} absolute prediction error and the (absolute and normalised) gradients of the data and \textbf{b)} the prediction error and predictive uncertainty (determined as the 1 $\sigma$ interval of DeepEnsemble predictions) of each surrogate.}
  \label{fig:heatmaps}
\end{figure}

\autoref{fig:heatmaps} displays two kinds of heatmaps for each surrogate to provide further insights into their correlational behaviour. The heatmaps are to be understood as a two-dimensional histograms with 100 x 100 bins. Notably, the colormap is applied on a log-scale due to the large disparities between dense and empty cells. The y-axes in both plots display the absolute prediction error. 

\autoref{fig:gradients_heatmap} displays the correlation between gradient magnitude and prediction errors, hence the x-axis shows the normalised absolute gradient. Overall, \verb|LatentNeuralODE| and \verb|LatentPoly| show broader scattering than \verb|FullyConnected| and \verb|MultiONet|, indicative of higher prediction errors for points with higher gradients. 

\autoref{fig:uq_heatmap} shows the correlation between the predicted uncertainty (using DeepEnsemble) and prediction errors, hence the x-axis is the predictive uncertainty. A perfect uncertainty quantification method would always predict the actual prediction error, signified here by the white diagonal line. Notably, the overconfidence of \verb|FullyConnected| discussed in \autoref{results} is clearly visible in the corresponding heatmap. The autoencoder-type models again exhibit broader scattering towards larger prediction errors and predictive uncertainties which is related to their lower overall accuracy. But for UQ, quantifying the PE is more relevant than achieving a low PE, and the autoencoder models seem to do this more reliably. It is worth noting that a large fraction of the total counts falls into the cell in the lower-left corner for all surrogates, as many of the PEs and corresponding uncertainties are too small to resolve.

\end{document}